\documentclass[a4paper,twoside]{article}

\usepackage{subcaption}
\usepackage{calc}
\usepackage{amssymb}
\usepackage{amstext}
\usepackage{amsmath}
\usepackage{amsthm}
\usepackage{multicol}
\usepackage{pslatex}
\usepackage{apalike}
\usepackage[ruled,vlined,linesnumbered]{algorithm2e}
\usepackage[bottom]{footmisc}
\usepackage{hyperref}
\usepackage[dvipsnames]{xcolor}
\usepackage{paralist}
\usepackage{multirow}
\usepackage{array}
\usepackage{booktabs}
\usepackage{placeins}
\usepackage{orcidlink}

\usepackage{SCITEPRESS}     
\SetAlFnt{\small}
\SetAlCapFnt{\small}
\SetAlCapNameFnt{\small}
\SetKwComment{Comment}{ $\triangleright$\ }{}%
\SetCommentSty{footnotesize}
\SetKw{KwEach}{each}
\SetKw{KwIn}{Input}
\SetKw{KwOut}{Output}
\SetKw{KwSplit}{Split}
\SetKw{KwSample}{Sample}
\SetKw{KwUpdate}{Update}
\SetKw{KwGenerate}{Generate}
\SetKw{KwShuffle}{Shuffle}
\usepackage{algorithmic}
\algsetup{linenosize=\tiny}

\DeclareMathOperator*{\argmin}{arg\,min}

\newcommand{\newalg}[0]{TX-Gen}

\newtheorem{definition}{Definition}
\newtheorem{assumption}{Assumption}
\begin{document}

\title{TX-Gen: Multi-Objective Optimization for Sparse Counterfactual Explanations for Time-Series Classification}

\author{\authorname{Qi Huang\sup{1}\orcidlink{0009-0007-4989-135X}, Sofoklis Kitharidis\sup{1}\orcidlink{0009-0005-8404-0724}, Thomas B{\"a}ck\sup{1}\orcidlink{0000-0001-6768-1478} and Niki van Stein\sup{1}\orcidlink{0000-0002-0013-7969}}
\affiliation{\sup{1}Institute of Advanced Computer Science, Leiden University, Einsteinweg 55, Leiden, The Netherlands}
\email{\{q.huang, s.kitharidis, t.h.w.baeck, n.van.stein\}@liacs.leidenuniv.nl}
}



\abstract{In time-series classification, understanding model decisions is crucial for their application in high-stakes domains such as healthcare and finance. Counterfactual explanations, which provide insights by presenting alternative inputs that change model predictions, offer a promising solution. However, existing methods for generating counterfactual explanations for time-series data often struggle with balancing key objectives like proximity, sparsity, and validity. In this paper, we introduce \newalg, a novel algorithm for generating counterfactual explanations based on the Non-dominated Sorting Genetic Algorithm II (NSGA-II). \newalg\ leverages evolutionary multi-objective optimization to find a diverse set of counterfactuals that are both sparse and valid, while maintaining minimal dissimilarity to the original time series. By incorporating a flexible reference-guided mechanism, our method improves the plausibility and interpretability of the counterfactuals without relying on predefined assumptions. Extensive experiments on benchmark datasets demonstrate that \newalg\ outperforms existing methods in generating high-quality counterfactuals, making time-series models more transparent and interpretable.}

\onecolumn \maketitle \normalsize \vfill

\section{\uppercase{Introduction}}
\label{sec:introduction}

The increasing adoption of machine learning models for time-series classification (TSC) in critical domains such as healthcare~\cite{morid2023time} and finance~\cite{chen2016financial} has raised the demand for interpretable and transparent decision-making processes. However, the black-box nature of many high-performing classifiers, such as neural networks or ensemble models, limits their interpretability, making it difficult for practitioners to understand why a particular decision is made. In this context, counterfactual explanations have emerged as a valuable approach in Explainable AI (XAI)~\cite{theissler2022explainable}, providing instance-specific insights by identifying alternative inputs that lead to different classification outcomes. Despite the growing body of work in this area, generating meaningful counterfactuals for time-series data presents unique challenges due to its sequential nature, dependency on temporal structure, and multi-dimensional complexity.

While counterfactual generation has been extensively explored for tabular and image data, methods tailored to time-series classification remain scarce and underdeveloped. The few existing methods often fail to balance key properties such as proximity, sparsity, and validity. Moreover, most approaches require high computational resources or rely on rigid heuristics, limiting their applicability in real-world scenarios. This gap motivates the development of an efficient, model-agnostic method capable of generating high-quality counterfactual explanations for time-series classifiers.

In this paper, we propose \newalg, a novel algorithm for generating counterfactual explanations in time-series classification tasks using a modified Non-dominated Sorting Genetic Algorithm II (NSGA-II). Unlike previous approaches that combine evolutionary computing with explainable AI \cite{zhou2024evolutionarycomputationexplainableai}, \newalg\  leverages the power of evolutionary multi-objective optimization to find a diverse set of Pareto-optimal counterfactual solutions that simultaneously minimize multiple objectives, such as dissimilarity to the original time-series and sparsity of changes, while ensuring the classifier’s decision is altered. Additionally, our approach incorporates a flexible reference-based mechanism, which guides the counterfactual generation process without relying on restrictive assumptions or predefined shapelets. 

\paragraph{Contributions}
The key contributions of this work are as follows:
\begin{itemize}
    \item We introduce \newalg, a novel framework for generating counterfactual explanations specifically tailored to time-series classification, which efficiently balances proximity, sparsity, and validity through multi-objective optimization.
    \item Our method employs a reference-guided approach to select informative subsequences, improving the plausibility and interpretability of counterfactuals.
    \item We demonstrate the effectiveness of \newalg\ across multiple benchmark datasets, showing that our approach outperforms existing methods in terms of sparsity, validity and proximity of the counterfactual examples.
\end{itemize}

In the following sections, we detail the methodology behind \newalg, present our experimental results, and discuss the implications of our findings for advancing explainability in time-series classification.




\subsection{Problem Statement}
\label{subsec: problem statement}
We begin our problem formulation by first introducing the context of this research. 

Given a collection (set) of univariate time series, denoted as \( S = \{T_1, T_2, \dots, T_n\} \), where each \(T = \{t_i \in \mathbb{R}\}_{i=1}^m \) consists of observations orderly recorded across $m$  timestamps. In the context of a time series classification (TSC) task, each time series $T$ is associated with a true descriptive label $L \in\{L_1,\ldots,L_k\}$ from $k$ unique categories. The objective of TSC is to develop an algorithmic predictor $f(\cdot)$ that can maximize the probability  \(P(f(T)=L) \) for $T\in S$ and also any unseen $T \notin S$ (provided that the true label of such a $T \notin S$ also equals $L$). 
Here, we presume the classifier provides probabilistic or logit outputs. Moreover, assume a function $g: \mathbb{R}^m \to \mathbb{R}^m$ that can element-wise perturb or transform any given time series $T$ into a new \textit{in-distribution} time series instance $\Hat{T}$. We then say that $\Hat{T}$ is a counterfactual of $T$ regarding a TSC predictor $f$ if \(f(\Hat{T})\neq f(T)\) while the dissimilarity between $T$ and $\Hat{T}$ is \textit{minimized}~\cite{wachter_counterfactual_2017,delaneyInstanceBasedCounterfactualExplanations2021a}. It is noteworthy that counterfactual explanations are fundamentally instance-based explanations aimed at interpreting the predictions of the classifier being explained, rather than the true labels, regardless of whether the real labels are known or not.

\paragraph{Goal} Considering a robust time series classification predictor \( f \) trained on the dataset \( S \), and a time series instance \( T_e \) to be explained, this work proposes an explainable AI method, i.e., the aforementioned function $g$, which can identify the counterfactual of \( T_e \) concerning \( f \). Notably, \( T_e \) may be either part of the dataset \( S \) or external to it.

\section{\uppercase{Background}}

\subsection{Evaluation Metrics}
The evaluation criteria for our counterfactual generation method are articulated through several key metrics, each addressing specific aspects of counterfactual quality and effectiveness:
\begin{itemize}
    \item \textbf{Proximity}: Measures the element-wise similarity between the counterfactual instance and the target time series, typically by using the  \( L_p \) metrics.
    
    \item \textbf{Validity}: Assesses whether the counterfactuals successfully alter the original class label, thereby reflecting their effectiveness.
    
    \item \textbf{Diversity}: Measures the variety in the generated counterfactual solutions for each to-be-explained time series instance.
    \item \textbf{Sparsity}: Quantifies the simplicity of the counterfactual by counting the number of element-wise differences between the generated counterfactual and the original series.
\end{itemize}
These metrics collectively provide a robust framework for evaluating the quality and utility of counterfactual explanations, emphasizing minimalism, diversity, validity, and closeness to the original instances.

\subsection{Related Work}
The evolution of explainable artificial intelligence (XAI) for generating counterfactuals for time series has been marked by significant advances in methods that provide clear, actionable insights into model decisions. This line of research has been initiated by~\cite{wachter_counterfactual_2017}, introducing an optimization-based approach that focuses on minimizing a loss function to modify decision outcomes while maintaining minimal Manhattan distance from the original input. This foundational method encouraged further development of algorithms that enhance the quality and effectiveness of counterfactual explanations through additional modifications to the loss function.

Building upon the concept of influencing specific features within time series data, the introduction of local tweaking via a Random Shapelet Forest classifier~\cite{karlsson_locally_2020} targets specific, influential shapelets for localized transformations within time series data.  In contrast, global tweaking~\cite{karlsson_explainable_2018} uses a k-nearest neighbor classifier to guide broader transformations ensuring minimal yet meaningful alterations to change class outcomes.

Advancing these concepts, \cite{delaneyInstanceBasedCounterfactualExplanations2021a}   developed the Native Guide Counterfactual Explanation (NG-CF)  which utilizes the Dynamic Barycenter Averaging with the nearest unlike neighbor. Subsequently, ~\cite{liSGCFShapeletGuidedCounterfactual2022} introduced the Shapelet-Guided Counterfactual Explanation (SG-CF), exploiting time series subsequences that are maximally representative of a class to guide the perturbations needed for generating counterfactual explanations. This method was further elaborated upon with the introduction of the Attention-Based Counterfactual Explanation for Multivariate Time Series (AB-CF) by~\cite{liAttentionBasedCounterfactualExplanation2023a} which incorporates attention mechanisms to further refine the selection of shapelets and optimize the perturbations. 

Moreover, TSEvo by~\cite{holligTSEvoEvolutionaryCounterfactual2022} utilizes the Non-Dominated Sorting Genetic Algorithm to strategically balance multiple explanation objectives such as proximity, sparsity, and plausibility while incorporating three distinct mutations. Most recently, Sub-SpaCE ~\cite{refoyoSubSpaCESubsequenceBasedSparse2024} employs genetic algorithms too, with customized mutation and initialization processes, promoting changes in only a select few subsequences. This method focuses on generating counterfactual explanations that are both sparse and plausible without extensive alterations.

Despite the progress in generating counterfactual explanations for time-series classification, these existing approaches still face several significant limitations. Many methods, such as optimization-based or shapelet-guided approaches, rely heavily on predefined assumptions about the data, such as the importance of specific subsequences or the necessity for local perturbations. These assumptions often reduce the generalizability of the methods across diverse datasets and classifiers. Furthermore, most methods struggle to balance the trade-offs between proximity, sparsity, and diversity of the generated counterfactuals, often prioritizing one metric at the expense of others. This can lead to counterfactuals that are either too similar to the original instance to be meaningful or excessively altered, making them unrealistic or hard to interpret. Additionally, many existing approaches lack scalability, becoming computationally expensive when applied to larger or more complex time-series data. In contrast, our proposed method, \newalg, addresses these limitations by leveraging the flexibility of evolutionary multi-objective optimization to dynamically explore the solution space and by guiding the search using reference samples from the training data. This allows \newalg\ to generate diverse, sparse, and valid counterfactuals with greater computational efficiency, making it more applicable to real-world time-series classification tasks.

\section{\uppercase{Methodology}}

\paragraph{The framework}
Our proposed method, \newalg, aims to jointly locate the subsequence of interest in a to-be-explained time series \( T \) while also transforming \( T \) into its counterfactual. This is achieved by mutually minimizing two objective functions under the guidance of reference samples, using a customized iterative optimization heuristic based on the Non-Dominated Sorting Genetic Algorithm II (NSGA-II)~\cite{debFastElitistMultiobjective2002}. A high-level pseudocode of our algorithm is given in Algorithm~\ref{alg:main} assuming the respect readers are familiar with the concepts of evolutionary algorithms~\cite{backHandbookEvolutionaryComputation1997,back2023evolutionary}
\begin{algorithm}[ht]
\SetAlgoLined
\caption{High-Level pseudocode for our counterfactual discovery algorithm \newalg\ \label{alg:main}}
\KwData{Population size $N$; crossover probability $p_c$; mutation probability $p_m$, number of generations $G$; to-be-explained time series $T$; classifier $f$; a set of references $S$; number of final reference instances $K$}
\KwResult{Set of non-dominated counterfactual candidate solutions}

Initialize population $P_0$ of size $N$ as described in section~\ref{subsec: chromosome}\\
\Comment*[f]{Selection of references using Algorithm~\ref{alg:choose reference}}\\
\(\psi \gets \texttt{SelectReference}(T, S, f, K)\)\\
Generate the counterfactual candidate sets $\Theta_0 = \bigcup_{i=1}^{N}\{\Hat{T}_i\}$ for all individuals in $P_0$ using the \texttt{Generate} function from Algorithm~\ref{alg:generation SoI}. \\
Evaluate the fitness of each individual in $P_0$ ($\Theta_0$) using two objective functions (see section~\ref{subsec: obj vals})\\
\For{ $t = 1$ to $G$}{
    Generate offspring $Q_t$ from $P_{t-1}$ by: \\
    - Select the parents for mating using binary tournament selection\\
    - \texttt{Crossover} the parents (with probability $p_c$) to obtain $Q'_{t}$ using Algorithm~\ref{alg:crossover} \\
    - \texttt{Mutate} the individuals in $Q'_{t}$ in place (with probability $p_m$) using Algorithm~\ref{alg:mutation}\\
    - Enumeratively expand each \(\overrightarrow{X_i} \in Q'_{t}\) into $K$ chromosomes and merge them into a new set
    \[Q_{t} = \bigcup_{i=1}^{2N} \bigcup_{j=1}^{K}\overrightarrow{X_{i, j}},\] where each \(\overrightarrow{X_{i, j}}=(x_{i,1}, x_{i, 2}, j)\)\\
    Obtain the counterfactual candidates $\Theta_t$ for all samples in $Q_{t}$ using \texttt{Generate} function\\
    
    Evaluate the fitness of each candidate in $\Theta_t$ and assign the resulting values to respective individuals in $Q_{t}$\\
    
    Combine parent population $P_{t-1}$ and offspring population $Q_t$ into a combined population $R_t$ of size $(2K+1)N$\\
    
    Perform non-dominated sorting on $R_t$ and assign crowding distances to individuals within each front\\
    
    Select individuals for the next population $P_t$ based on rank and crowding distance, ensuring the population size is $N$\\
}

Return the non-dominated individuals as well as their corresponding counterfactual candidates from the final population $P_G$
\end{algorithm}
Following the self-explainable high-level pseudocode, the key components proposed in our algorithm are further explained. 

\subsection{Model-based Selection of References}
\label{subsec: reference selection}
In TSC, consider a time series instance  T  that belongs to category $C$. A counterfactual-based explainer can be viewed as a reference-guided method if it leverages example instances—originating from the same task but known to belong to categories other than  $C$ —to assist in \textit{transforming} T into its counterfactual. Existing literature on reference-guided methods primarily identifies the reference instances by selecting those with low shape-based or distance-based similarities to the given instance to be explained, e.g., the nearest unlike neighbors as seen in~\cite{delaneyInstanceBasedCounterfactualExplanations2021a,holligTSEvoEvolutionaryCounterfactual2022}. In contrast, our work proposes selecting reference samples solely based on the classifier’s outputs. 

Following the settings in Section~\ref{subsec: problem statement}, let \( f(\cdot) \) be a probabilistic classifier for a TSC task \( \Omega \) with \( k \) classes. That is, instead of predicting the exact label, \( f \) predicts the probability distribution over all candidate class labels for any given target time series \( T \).

\begin{definition}[Distance in the Classifier]
\label{def:dist in classifier}
For any pairs of time series \((T_i, T_j)\) of the TSC task $\Omega$, we define their Distance in the Classifier $f$ as the Jensen-Shannon distance between \(f(T_i)\) and \(f(T_j)\):
\begin{equation}
    D_f(T_i, T_j) = \sqrt{\frac{KL(f(T_i) \mid P_m) + KL(f(T_j) \mid P_m)}{2}},
\end{equation}
where \(P_m\) is the element-wise mean of \(f(T_i)\) and \(f(T_j)\), and $KL$ denotes the well-known Kullback-Leibler (KL) divergence~\cite{endresNewMetricProbability2003}. Apart from the properties that a valid metric holds, this distance is also bounded, \(D_f(T_i, T_j) \in [0, 1]\) if we use 2 as the logarithm base in KL divergence.
\end{definition}

With this definition, as described in Algorithm~\ref{alg:choose reference}, we can retrieve reference instances for a to-be-explained time series \( T \) from a candidate set that contains samples that both are predicted to belong to different categories than \( T \) and are close to \( T \) with respect to the distance in the classifier.

\begin{algorithm}[htbp]

\caption{Select References for a Time Series.\\
\textbf{Input:} Target time series $T$; the classifier $f$; reference set $S$; number of reference samples $K$\\
\textbf{Output:} A set of reference samples \(\psi \subseteq S\).\label{alg:choose reference}}

\SetAlgoLined
\DontPrintSemicolon
  \SetKwFunction{FMain}{SelectReference}
  \SetKwProg{Pn}{Function}{:}{}
  \Pn{
  \FMain{$T$, $S$, $f$, $K$}}
  {
    
    \ForEach{\(T_i \in S\)}{
    \uIf{\(f(T_i)\neq f(T)\)}{
        Compute \(D_f(T, T_i)\) as in Definition~\ref{def:dist in classifier}\;
    }
    \Else{
    Set \(D_f(T, T_i) \gets 1.01\)
    }
    }
    Get the set: \(\psi \gets \argmin_{\substack{T_i \in S\\ |\psi|=K}} D_f(T, T_i)\)

}
\KwRet $\psi$
\end{algorithm}

\subsection{The Objective Functions}
\label{subsec: obj vals}
To address the idea (see section~\ref{subsec: problem statement}) that a decent counterfactual \(\Hat{T} = \{\Hat{t_i} \in \mathbb{R}\}_{i=1}^m \) shall closely resemble the target time series \(T = \{t_i \in \mathbb{R}\}_{i=1}^m \) while flipping the prediction of a given classifier $f$, we propose to determine feasible candidates \(\Hat{T}^{*}\) by jointly \textbf{minimizing} two real-valued objectives as follow:
\begin{itemize}
    \item The minimum Distance in Classifier between a counterfactual candidate $\Hat{T}^{*}$ and the samples in the reference set $\psi$:
    \begin{multline*}
    F_1(\Hat{T}^{*}, \psi) = \\
    \begin{cases}
    \min D_f(\Hat{T}^{*}, T_i),\ \forall T_{i} \in \psi & \text{if } f(\Hat{T}^{*}) \neq f(T), \\
    1.01 & \text{if } f(\Hat{T}^{*}) = f(T).
    \end{cases}
    \end{multline*}
    The second condition is to penalize the case where \(\Hat{T}^{*}\) fails to flip the prediction of the classifier.
    \item Joint sparsity and proximity between \(\Hat{T}^{*}\) and \(T\):
    \begin{align*}
        F_{2,1}(\Hat{T}^{*}, T) &= \frac{\sum_{
        i = [1,\dots,m]
        } \mathbf{1}_{\Hat{t_i}\neq t_i}}{m},\\
        F_{2,2}(\Hat{T}^{*}, T) &= \frac{|\Hat{T}^{*} - T|_2}{|\Hat{T}^{*}|_2 + |T|_2},\\
        F_2(\Hat{T}^{*}, T) &= \frac{1}{2} (F_{2,1}(\Hat{T}^{*}, T) + F_{2,2}(\Hat{T}^{*}, T))
    \end{align*}
    Here we normalize and combine the two indicators into one objective (\(F_2\)), where the sparsity (\(F_{2,1}\)) measures the degree of element-wise perturbation, and the second indicator \(F_{2,2}\) aims to quantify the scale change between \(\Hat{T}^{*}\) and \(T\).
\end{itemize}
In summary, the two objective functions can jointly ensure the found counterfactual solutions are valid and minimized. What is more, the proposed two objective functions are both bounded, saying \(F_1 \ \in [0,1]\) and \(F_2 \in [0,1]\).

\subsection{Locating the Sequences of Interest}
\subsubsection{Chromosome Representation}
\label{subsec: chromosome}
Previous works that utilize evolutionary computation predominantly adopt a unified chromosome representation that encodes the entire counterfactual instance. In other words, the number of variables in each solution is at least as large as the number of timestamps in the target time series. Furthermore, these approaches typically apply crossover, mutation, and other evolutionary operators directly to the chromosomes to generate the next set of feasible candidates in one-go~\cite{holligTSEvoEvolutionaryCounterfactual2022,refoyoSubSpaCESubsequenceBasedSparse2024}. One advantage of encoding the entire time series into the individual is that it can find multiple \textbf{Segment or Subsequence of Interest (SoI)} at the same time.

In contrast, our approach reduces the complexity of candidate solutions by representing each one with only three mutable integer variables, i.e., \(X = [x_1, x_2, x_3]\), regardless of the number of timestamps in the target time series. Specifically,  \(x_1\)  and  \(x_2\)  denote the starting and ending indices of the single subsequence of interest considered in this individual, respectively. Meanwhile,  \(x_3\)  represents the index of the reference sample in the set \(\psi\), which is used to guide the discovery of potential counterfactual observations for the SoI (see section~\ref{subsec:predict the observations}).

Similar to previous approaches, our method can also identify multiple SoI for each target time series by generating various counterfactual solutions. Furthermore, by leveraging the concept of Pareto efficiency in multi-objective optimization, our approach ensures that the diverse SoI discovered are all equally significant.
\paragraph{Initialization} Several existing works explicitly incorporate low-level XAI strategies to identify the SoI in the target instance, thereby facilitating the counterfactual search process. These strategies include feature attribution methods, such as the GradCAM family~\cite{selvarajuGradCAMVisualExplanations2017}, which are employed in works like~\cite{delaneyInstanceBasedCounterfactualExplanations2021a,refoyoSubSpaCESubsequenceBasedSparse2024}. Additionally, subsequence mining methods, such as Shapelet extraction~\cite{yeTimeSeriesShapelets2009}, are utilized in studies like~\cite{liSGCFShapeletGuidedCounterfactual2022,huangShapeletbasedModelagnosticCounterfactual2024}. Unlike other XAI techniques, we do not use this (possibly biased) initialization, instead the first population in our search algorithm is randomly sampled. Given a to-be-explained time series \(T = \{t_i \in \mathbb{R}\}_{i=1}^m \) and its reference sample set $\psi$ of size $K$, the variables in a individual solution \(\overrightarrow{X_j}=[x_{j,1}, x_{j,2}, x_{j,3}]\) are step-by-step uniformly randomly sampled as:
\begin{align*}
    x_{j,1} &\sim \mathbb{U}(1, m-1) \\
    x_{j,2} &\sim \mathbb{U}(x_{j,1},m)\\
    x_{j,3} &\sim \mathbb{U}(1,K)
\end{align*}
\subsubsection{Crossover}
Crossover is a crucial operator in evolutionary algorithms, designed to create new offspring by combining decision variables from parent solutions. This process helps the algorithm exploit the search space within the current population. The parent solutions are chosen through a \textit{selection} mechanism. In this work, we employ the well-known tournament selection method to select mated parents, and we refer interested readers to the literature~\cite{goldbergComparativeAnalysisSelection1991} for more details.

Bearing in mind the idea behind crossover, our customized crossover operator is dedicated to fulfilling two additional criteria: \begin{inparaenum}[(i)]
    \item minimizing the intersection between the SoI of offspring.
    \item minimizing the total lengths of the SoI tracked by offspring.
\end{inparaenum}

\begin{algorithm}[htbp]

\caption{The Crossover Operator.\\
\noindent\textbf{Input:}  The parent individuals $\overrightarrow{X_1}=(x_{1,1}, x_{1,2}, x_{1,3})$ and $\overrightarrow{X_2}=(x_{2,1}, x_{2,2}, x_{2,3})$; crossover probability $P_{cx}$. \\
\textbf{Output:} The two offspring $\overrightarrow{Y_1} = (y_{1,1}, y_{1,2}, y_{1,3})$ and $\overrightarrow{Y_2} = (y_{2,1}, y_{2,2}, y_{2,3})$.
\label{alg:crossover}}

\SetAlgoLined
\DontPrintSemicolon
  \SetKwFunction{FMain}{Crossover}
  \SetKwFunction{FUtil}{GetUnique}
  \SetKwProg{Pn}{Function}{:}{}
    \Pn{
  \FUtil{$a_1, a_2, a_3, a_4$}}
  {
  \(b_1, b_2, b_3, b_4 \gets \text{AscendingSort}(a_1, a_2, a_3, a_4)\)\;
  \(\overrightarrow{\beta} \gets (b_1) \)\;
  \If{\(b2 \neq b1\)}{
  \(\overrightarrow{\beta} \gets \overrightarrow{\beta} \mathbin\Vert (b_2)\) \Comment*[r]{Concatenate $b_2$ to $\overrightarrow{\beta}$}
  }
  \If{\(b3 \neq b2\)}{
  \(\overrightarrow{\beta} \gets \overrightarrow{\beta} \mathbin\Vert (b_3)\)\;
  }
  \If{\(b4 \neq b3\)}{
  \(\overrightarrow{\beta} \gets \overrightarrow{\beta} \mathbin\Vert (b_4)\)\;
  }
  }
  \KwRet $\overrightarrow{\beta}$\;
  \Pn{
  \FMain{$\overrightarrow{X_1}$, $\overrightarrow{X_2}, P_{cx}$}}
  { 
    \(p_{c} \sim \mathbb{U}(0,1)\)\;
    \(\overrightarrow{\alpha} \gets \FUtil(x_{1,1}, x_{1,2}, x_{2,1}, x_{2,2})\)\;
        \(\overrightarrow{Y_1}, \overrightarrow{Y_2} \gets \overrightarrow{X_1}, \overrightarrow{X_2}\)\;
    \If{\(p_{c} \leq P_{cx} \)}{
        \uIf{
        \(|\overrightarrow{\alpha}| = 4\)
        }{
        \(A \gets |x_{1,1} - x_{2,1}| + |x_{1,2} - x_{2,2}|\)\;
        \(B \gets |x_{1,1} - x_{2,2}| + |x_{1,2} - x_{2,1}|\)\;
        \uIf{$A \leq B$}{
        \(y_{1,1} \gets \min (x_{1,1}, x_{2,1})\)\;
        \(y_{1,2} \gets \max (x_{1,1}, x_{2,1})\)\;
        \(y_{2,1} \gets \min (x_{1,2}, x_{2,2})\)\;
        \(y_{2,2} \gets \max (x_{1,2}, x_{2,2})\)\;
        }
        \Else{
        \(y_{1,1} \gets \min (x_{1,1}, x_{2,2})\)\;
        \(y_{1,2} \gets \max (x_{1,1}, x_{2,2})\)\;
        \(y_{2,1} \gets \min (x_{1,2}, x_{2,1})\)\;
        \(y_{2,2} \gets \max (x_{1,2}, x_{2,1})\)\;
        }
        }
        \uElseIf{
        \( |\overrightarrow{\alpha}| = 3\)}{
            \(y_{1,1}, y_{1,2} \gets \alpha_1, \alpha_2\)\;
            \(y_{2,1}, y_{2,2} \gets \alpha_2, \alpha_3\)\;
        }
        \Else{
            \(y_{1,1}, y_{2,2} \gets \alpha_1, \alpha_2\)\;
            \(y_{1,2} \sim \mathbb{U}(\alpha_1, \alpha_2)\)\;
            \(y_{2,1} \gets y_{1,2}\)\;
        }
    }
}
\KwRet ($\overrightarrow{Y_1}$, $\overrightarrow{Y_2}$)
\end{algorithm}

The proposed crossover is demonstrated in Algorithm~\ref{alg:crossover}. The crossover operation by default produces two offspring. The function \FUtil returns the unique, non-duplicate values of its inputs in an array, sorted in ascending order. Notably, in line 22 of the pseudocode, we compare the total lengths of the SoI under two feasible crossover options and intentionally set the relational operator to \( \leq \) instead of \( < \). This ensures that, in cases where the SoIs of two parents do not overlap, their offspring will have minimal intersections. Between lines 33 and 40, we address the corner cases where the indices recorded in the parents are partially or entirely the same.
 
\subsubsection{Mutation}
In general, the mutation operators help to explore unforeseen areas of the solution space and prevent premature convergence. Unlike the crossover operator that mixes the SoIs from different parent solutions, the mutation operator proposed in this work aims to adjust the length of the SoI tracked by each individual selected for mutation. The pseudocode of our mutation strategy is provided in Algorithm~\ref{alg:mutation}.

\begin{algorithm}[htbp]

\caption{The Mutation Operator.\\
\textbf{Input:}  The individual $\overrightarrow{X}=(x_{1}, x_{2}, x_{3})$; mutation probability $P_{mu}$; tolerable SoI length ratio \(\tau\); \(M\), the number of timestamps (length) of the target time series \(T\). \\
\textbf{Output:} The offspring $\overrightarrow{Y} = (y_{1}, y_{2}, y_{3})$ after mutation.
\label{alg:mutation}}

\SetAlgoLined
\DontPrintSemicolon
  \SetKwFunction{FMain}{Mutate}
  \SetKwProg{Pn}{Function}{:}{}
  \Pn{
  \FMain{$\overrightarrow{X}, P_{mu}, \tau, M$}}
  { 
      \(\overrightarrow{Y} \gets \overrightarrow{X}\)\;
    \(p_{u} \sim \mathbb{U}(0,1)\)\;
    \Comment*[f]{Configure the $p$ of a \(\textbf{Bin}(n, p)\)}\;
    \If{\(p_{u} \leq P_{mu} \)}{
    \uIf{$\tau \in (0, 1)$}{
    \(\sigma \gets \frac{\log(0.5)}{\tau} \cdot \frac{x_{2}-x_{1}}{M}\)\;
    \(p_b \gets e^{\sigma} \)\;
    }
    \Else{
    \(p_b \gets 0.5 \)
    }
    \Comment*[f]{Determine the direction to scale}
    \(p_s \sim \mathbb{U}(0,1)\)\;
    \Comment*[f]{Sample the new SoI length}
    \( l \sim \textbf{Bin}(2 (x_2 - x_1), p_b)\)\;
    \uIf{\(p_s \leq 0.5 \)}{
    \(y_2 \gets \min (\max (y_1+l, y_1 + 1), M)\)
    }
    \Else{
    \(y_1 \gets \max (\min (y_2 - l, y_2 - 1), 1)\)
    }
        
    }
}
\KwRet \(\overrightarrow{Y}\)
\end{algorithm}

\begin{figure}[htbp]
    \centering
    \begin{subfigure}{\columnwidth}
    \centering
    \includegraphics[width=\linewidth]{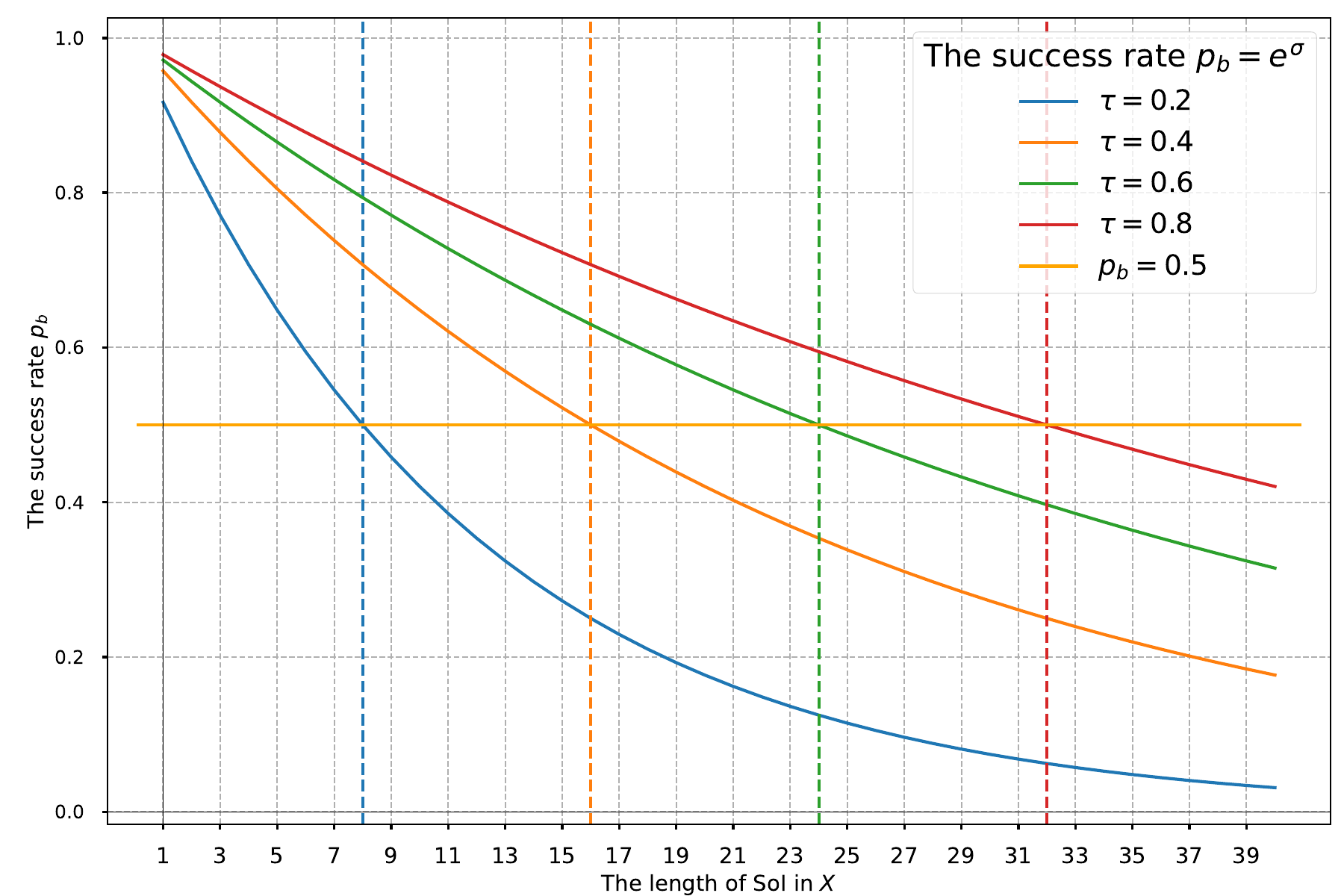}
    \caption{An empirical plot of multiple success rates $p_b$ versus the length of SoI in to-be-mutated individuals $X$ under various tolerable SoI length ratios $\tau$.}
    \label{fig:example binomial p}
    \end{subfigure}\\
    \vspace{0.5cm}
    \begin{subfigure}{\columnwidth}
        \includegraphics[width=\linewidth]{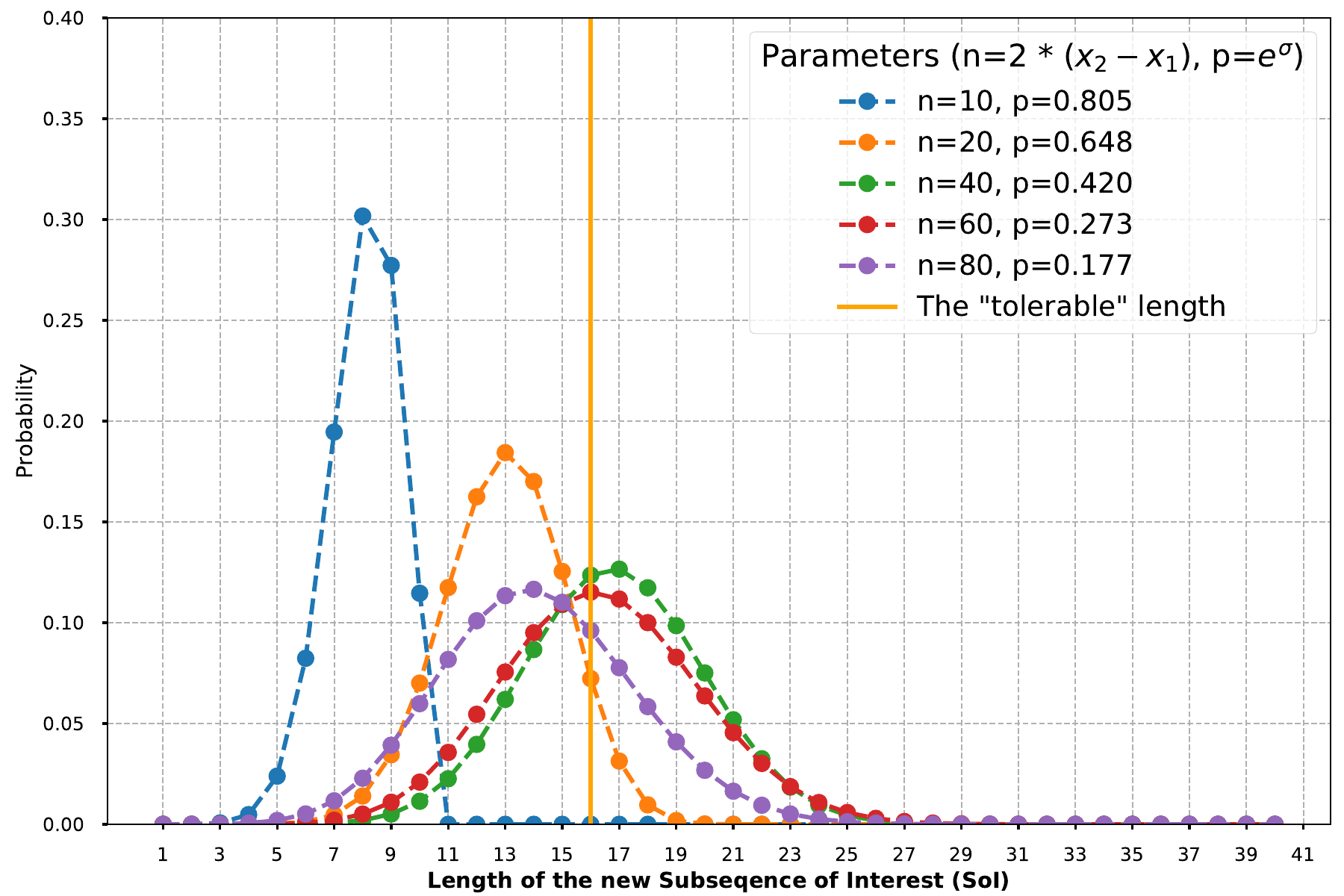}
        \caption{An example Probability Mass Function (PMF) of a binomial distribution under our configuration. The tolerable SoI length ratio is set to $\tau = 0.4$, which results in a tolerable length of 16. And the five displayed PMFs are calculated based on SoI lengths 5, 10, 20, 30, and 40, respectively.}
        \label{fig:example binomial PMF}
    \end{subfigure}
    
    \caption{ In both figures, the length of the to-be-explained time series is $M=40$.}
     \label{fig:example binomial}
\end{figure}

The strategy employs a dynamic scaling of the SoI length for the new instance \( \overrightarrow{Y} \) based on the SoI length of the previous instance \( \overrightarrow{X} \). The new length is randomly sampled from a binomial distribution \(\textbf{Bin}(n, p)\), where the number of trials \( n \) is twice the length of the SoI in \( \overrightarrow{X} \). This approach allows the SoI length to potentially increase almost exponentially between adjacent generations during optimization, effectively simulating the idea of a binary search. Another vital parameter to specify a binomial distribution is $p$, the success rate of trials, which can be interpreted as the probability of extending ($p > 0.5$) or shrinking ($p < 0.5$) the SoI in this context. In lines 5 to 10 of Algorithm~\ref{alg:mutation}, given \( \tau \), a tolerable (or ideal) ratio of the SoI length to the length of the time series, the value \( p \) is determined such that the peak of the probability mass function of the binomial distribution is close to \( \lfloor M \cdot \tau \rfloor \). Empirical illustrations of this concept are provided in Figure~\ref{fig:example binomial}. As shown in Figure~\ref{fig:example binomial p}, under the same \( \tau \), the success rate decreases (increases) as the SoI length increases (decreases). Additionally, for the same SoI length, an increase (decrease) in \( \tau \) leads to a corresponding increase (decrease) in the likelihood of extending the SoI. Furthermore, Figure~\ref{fig:example binomial PMF} demonstrates that the probability of sampling a new length close to the tolerable length is increased, enabling the adaptive adjustment of the SoI length, particularly discouraging excessively long SoIs. This can be observed as the \textit{peak} of \(n = 80\) (SoI length 40) is shifted further left of the tolerable length compared to that of \(n = 40\) (SoI length 20).

\subsection{Modeling of the New SoI}
\label{subsec:predict the observations}
It is noteworthy that several previous works also separate counterfactual generation into two stages: the discovery of SoIs (or other forms of sensitivity analysis) and the subsequent generation or search for the actual values. While the crossover and mutation operators described above facilitate the search for optimal SoIs within the time series, the specific values corresponding to these subsequences must still be determined to find true counterfactual explanations. Existing approaches to determining the values for SoIs can be broadly categorized into three types:
\begin{inparaenum}
    \item heuristic methods, which iteratively optimize specific criteria~\cite{wachter_counterfactual_2017,holligTSEvoEvolutionaryCounterfactual2022};
    \item direct use of (sub)sequences from reference data, typically selected from the training set~\cite{delaneyInstanceBasedCounterfactualExplanations2021a,liSGCFShapeletGuidedCounterfactual2022};
    \item generation of values using machine learning models, such as generative models~\cite{langGeneratingSparseCounterfactual2023,huangShapeletbasedModelagnosticCounterfactual2024}.
\end{inparaenum}
Interestingly, the second strategy essentially complements the last. By combining these two approaches, it becomes feasible to obtain an SoI efficiently without being restricted to values from the reference data, which is the foundation of our proposed approach. 

The proposed method relies on two assumptions. Given a univariate time series \(T\) of length $m$ and one of its counterfactual candidate \(\Hat{T}\). Suppose the located subsequence of interest is between timestamps $i$ and $j$ where $i < j$, then the first assumption is as follows:

\begin{assumption}
The time series \( T = \{t_k \in \mathbb{R}\}_{k=1}^m \) and its counterfactual \( \Hat{T} = \{\Hat{t_k} \in \mathbb{R}\}_{k=1}^m \) differ only in the subsequence of interest, i.e., \( t_k = \Hat{t_k}, \ \forall k \notin [i, j] \).
\end{assumption}
While this assumption aims to ensure the true importance of the found SoI, the second assumption is developed to further restrict the values of SoI to a simplified form for efficient modeling and generation.
\begin{assumption}
The difference between the SoI of the counterfactual candidate \( \Hat{T} = \{\Hat{t_k} \in \mathbb{R}\}_{k=1}^m \) and that of the target time series \( T = \{t_k \in \mathbb{R}\}_{k=1}^m \), denoted as \( Z = \{\Hat{t_k} - t_k\}_{k=i}^{j} \), is a stationary process where the current values are linearly dependent on its past values. Consequently, \( Z \) can be modeled by an autoregressive model of order \( p \) (AR-\( p \)) with a Gaussian error term.
\end{assumption}
This assumption is straightforward and aims to define the nature of the difference between the SoIs. To the best of our knowledge, in contrast to previous approaches that directly obtain the observations (values) of counterfactuals, we are the first to model the difference between the counterfactual and the to-be-explained time series instance. Moreover, in section~\ref{subsec: reference selection}, it is mentioned that this method is reference-guided. By utilizing the reference instances found using Algorithm~\ref{alg:choose reference} and the chromosome representations, we now describe the methodology of generating counterfactual instances in Algorithm~\ref{alg:generation SoI}.
\begin{algorithm}[htbp]
\caption{Generation of the SoI of counterfactual candidate\\
\textbf{Input:}  The target time series \( T = \{t_k \in \mathbb{R}\}_{k=1}^m \); the individual $\overrightarrow{X}=(x_{1}, x_{2}, x_{3})$; a reference sample $\widetilde{T}= \{\Tilde{t}_k \in \mathbb{R}\}_{k=1}^m$; the order (lags) of the autoregressive model $p$.\\
\textbf{Output:} The counterfactual candidate \(\Hat{T}\)
\label{alg:generation SoI}}

\SetAlgoLined
\DontPrintSemicolon
  \SetKwFunction{FMain}{Generate}
  \SetKwProg{Pn}{Function}{:}{}
  \Pn{
  \FMain{$T, \overrightarrow{X}, \psi, p, m$}}
  { 
    \(\Hat{T} \gets T\)\;
    \(l \gets x_2 - x_1\)\;
    \(i,\ j \gets \max (x_1 - p, 1),\ \min (x_2 + p, m)\)\;
    \(\zeta \gets \{\Tilde{t}_k - t_k\}_{k=i}^{j}\)\;
    
    Fit an autoregressive model $\texttt{AR}(p)$ on \(\zeta\)\;
    Predict \(\overline{\zeta} \gets \{\overline{t}_k\}_{k=1}^{j-i+1}\) using the $\texttt{AR}(p)$\;
    \(i^{*}  \gets \min(x_1, p)\)\;
    \(j^{*} \gets i^{*} + l\)\;
    
    \(\Hat{T}[x_1:x_2] \gets \overline{\zeta}[i^{*}:j^{*}] + T[x_1:x_2]\)\;
    }
\KwRet \(\Hat{T}\)
\end{algorithm}
Given a target time series \( T \) and a chromosome representation of the SoI \( \overrightarrow{X} \), the algorithm generates a counterfactual candidate based on a chosen instance $\Hat{T}$ from the reference set. In lines 4 and 5, the element-wise differences between the reference and the target are computed as a new time series \( \zeta \). Further, from lines 6 to 10, an autoregressive model of order \( p \) is fitted to learn the distribution of \( \zeta \) through conditional maximum likelihood estimation. The in-sample predictions are then added to the SoI values of \( T \) to form the counterfactual SoI. In this context, the autoregressive model essentially acts as a process of denoising and \textit{smoothing}.

\section{\uppercase{Experimental Setup}}
The experiments are performed using a diverse selection of time-series classification benchmarks and two popular time-series classifiers.

The UCR datasets ~\cite{bagnallGreatTimeSeries2017a} from the time-series classification archive~\cite{UCRArchive} chosen in this study represent a diverse array of application domains in time-series classification, all of which are one-dimensional. Furthermore, they are categorized in \hyperref[tab:datasets]{Table~\ref{tab:datasets}} based on their domain specificity and characteristics, providing a comprehensive assessment of the proposed method's robustness and adaptability across varied types of time series data.

    
    
    
    

\begin{table}[!hbtp]
\centering
\caption{The summary of the datasets along with the accuracy of the trained classifiers used in the experiments.}
\label{tab:datasets}
\resizebox{0.475\textwidth}{!}{
\begin{tabular}{c|cccccc}
\toprule
\textbf{Dataset} & \textbf{Length} & \textbf{Train + Test Size} & \textbf{Balanced} & \textbf{Classes} & \textbf{Catch22} & \textbf{STSF} \\ 
\midrule
\midrule
ECG200           & 96              & 100                 + 100                & No           & 2                & 0.83                & 0.87              \\ 
GunPoint         & 150             & 50                  + 150                & Yes           & 2                & 0.94                & 0.94              \\
Coffee           & 286             & 28                  + 28                 & Yes       & 2                & 1.00               & 0.964             \\
\midrule
CBF              & 128             & 30                  + 900                & Yes    & 3                & 0.96                & 0.98              \\ 
Beef             & 470             & 30                  + 30                 & Yes       & 5                & 0.60                & 0.66              \\
Lightning7       & 319             & 70                  + 73                 & No        & 7                & 0.69                & 0.75              \\ 
ACSF1            & 1460            & 100                 + 100                & Yes        & 10               & 0.87                & 0.82              \\ 
\bottomrule
\end{tabular}
}
\end{table}

For the classification tasks within our framework, each dataset was trained and tested using two specific classifiers: the \textit{Catch22} classifier (C22)~\cite{lubbaCatch22} and the \textit{Supervised Time Series Forest} (TSF)~\cite{SupervisedTSForest}. These classifiers were meticulously chosen not only for their computational efficiency and ease of implementation but also for their ability to produce probabilistic outputs. The generation of probabilistic outputs, rather than mere binary labels, is crucial for our methodology. It allows for the nuanced detection of subtle variations in the probability distributions across different classes. This feature is integral to our approach as it supports the generation of counterfactual explanations that are highly sensitive to minor but significant shifts in the data. 

\subsection{Hyperparameters Setup}
The customized NSGA-II algorithm central to our approach is designed with several hyperparameters that can influence its operation and performance.
The standard NSGA-II hyper-parameters (and their setting between brackets), population size (50), number of generations (50), the probability of crossover (0.7) and probability of mutation (0.7), are very common in evolutionary optimization algorithms, and are set based on the results of a few trials.
A new hyper-parameter, \textbf{Number of Reference Instances}, is set to 4 in this work. This parameter specifies both the number of reference cases used to evaluate the performance of a candidate solution and the number of \textit{teachers} employed to guide the search for finding counterfactual examples.
These hyper-parameters could be further optimized in future work. 


\subsection{Metrics and Baselines}
\label{subsec: evaluation metrics}
As introduced in section~\ref{subsec: evaluation metrics}, five criteria can be used to assess the effectiveness of the counterfactual explanation algorithms. Given a to-be-explained time series $T = \{t_i \in \mathbb{R}\}_{i=1}^m$, a set of $n$ counterfactuals candidates \(\Theta = \{\Hat{T_j} = \{\Hat{t_i} \in \mathbb{R}\}_{i=1}^m\}_{j=1}^{N} \), and the classifier $f$, we define the five evaluation metrics as follow:
\begin{enumerate}
    \item 
    \(
    \text{L1-Proximity}(T, \Hat{T}) = \frac{|\Hat{T} - T|}{|\Hat{T}| + |T|}
    \)
    \item 
    \(
    \text{L2-Proximity}(T, \Hat{T}) = \frac{|\Hat{T} - T|_2}{|\Hat{T}|_2 + |T|_2}
    \)
    \item \(
    \text{Validity}(T, \Hat{T} \mid f) = \mathbf{1}_{f(T)\neq f(\Hat{T})}
    \)
    \item 
    \(
    \text{Sparsity}(T, \Hat{T}) = \frac{\sum_{
        i = [1,\dots,m]
        } \mathbf{1}_{\Hat{t_i}\neq t_i}}{m}
    \)
    \item \(\text{Diversity}(T, \Theta \mid f) = \)\\
    \[
         \sum_{i=1}^{N-1} (\prod_{j=i+1}^{N} \mathbf{1}_{\Hat{T_i} \neq \Hat{T_j}}) \cdot \mathbf{1}_{f(T)\neq f(\Hat{T_i})},\ \Hat{T_i}, \Hat{T_j} \in \Theta
    \]
\end{enumerate}
Among the five metrics, the first four are pairwise metrics. In contrast, Diversity is focused on measuring the number of unique and valid counterfactuals generated by the explainer for each target time series in a single run. 
In this work, we evaluate the performance of \newalg\ against four well-established algorithms which can be separated into two groups: w-CF~\cite{wachter_counterfactual_2017} and Native-Guide~\cite{delaneyInstanceBasedCounterfactualExplanations2021a}, the hall-of-fame baselines; TSEvo~\cite{holligTSEvoEvolutionaryCounterfactual2022} and AB-CF~\cite{liAttentionBasedCounterfactualExplanation2023a}, the state-of-the-art counterfactual explainers.



\begin{table}[ht]
\centering
\resizebox{0.47\textwidth}{!}{
\begin{tabular}{l|*{8}{c@{\hskip 10pt}c}}
\toprule
\multirow{2}{*}{Explainer} & \multicolumn{1}{c}{ECG200} & \multicolumn{1}{c}{Coffee} & \multicolumn{1}{c}{GunPoint} & \multicolumn{1}{c}{CBF} & \multicolumn{1}{c}{Lightning7} & \multicolumn{1}{c}{ACSF1} & \multicolumn{1}{c}{Beef} \\
                           & Validity & Validity & Validity & Validity & Validity & Validity & Validity \\
\midrule
\midrule
NG (C22) & 0.56 & 0.46 & 0.43 & 0.59 & -- & -- & 0.83 \\
w-CF (C22) & 0.03 & -- & 0.01 & 0.01 & -- & 0.11 & 0.13 \\
\midrule
AB-CF (C22) & 0.93 & \textcolor{blue}{1.00} & \textcolor{blue}{1.00} & 0.99 & 0.99 & \textcolor{blue}{1.00} & \textcolor{blue}{1.00} \\
TSEvo (C22) & 0.39 & 0.54 & 0.45 & 0.50 & 0.56 & 0.51 & 0.07 \\
\midrule
\textbf{\newalg}  (C22) & \textcolor{blue}{1.00} & \textcolor{blue}{1.00} & \textcolor{blue}{1.00} & \textcolor{blue}{1.00} & \textcolor{blue}{1.00} & \textcolor{blue}{1.00} & \textcolor{blue}{1.00} \\
\midrule
\midrule
NG (TSF) & 0.35 & 0.50 & 0.50 & -- & -- & -- & 0.10 \\
w-CF (TSF) & -- & -- & -- & 0.01 & 0.11 & 0.17 \\
\midrule
AB-CF (TSF) & 0.74 & \textcolor{blue}{1.00} & \textcolor{blue}{1.00} & \textcolor{blue}{1.00} & 0.99 & 0.99 & 0.97 \\
TSEvo (TSF) & 0.35 & 0.50 & 0.50 & 0.35 & 0.44 & 0.22 & 0.20 \\
\midrule
\textbf{\newalg} (TSF) & \textcolor{blue}{1.00} & \textcolor{blue}{1.00} & \textcolor{blue}{1.00} & \textcolor{blue}{1.00} & \textcolor{blue}{1.00} & \textcolor{blue}{1.00} & \textcolor{blue}{1.00} \\
\bottomrule
\end{tabular}
}
\caption{The \textbf{Validity} metric for each counterfactual XAI method on two different classifiers (TSF and C22) and seven different benchmark datasets.\label{tab:validity}}
\end{table}

\begin{table*}[!htbp]
\centering
\resizebox{\textwidth}{!}{
\begin{tabular}{l|*{8}{c@{\hskip 10pt}c}}
\toprule
\multirow{2}{*}{Explainer} & \multicolumn{2}{c}{ECG200} & \multicolumn{2}{c}{Coffee} & \multicolumn{2}{c}{GunPoint} & \multicolumn{2}{c}{CBF} & \multicolumn{2}{c}{Lightning7} & \multicolumn{2}{c}{ACSF1} & \multicolumn{2}{c}{Beef} \\
                           & Mean & Std. & Mean & Std. & Mean & Std. & Mean & Std. & Mean & Std. & Mean & Std. & Mean & Std. \\
\midrule
NG (C22) & 0.403 & 0.310 & 0.962 & 0.131 & 0.052 & 0.065 & 0.989 & 0.067 & -- & -- & -- & -- & 1.000 & 0.000 \\
w-CF (C22) & \textcolor{blue}{0.010} & 0.000 & -- & -- & \textcolor{blue}{0.007} & 0.000 & 0.669 & 0.468 & -- & -- & 0.092 & 0.287 & 0.252 & 0.432\\
\midrule
AB-CF (C22) & 0.430 & 0.254 & 0.437 & 0.106 & 0.525 & 0.296 & 0.461 & 0.276 & 0.350 & 0.269 & 0.534 & 0.334 & 0.467 & 0.267 \\
TSEvo (C22) & 0.408 & 0.154 & 0.280 & 0.189 & 0.237 & 0.197 & 0.485 & 0.213 & 0.347 & 0.271 & 0.558 & 0.291 & 0.698 & 0.302 \\
\midrule
\textbf{\newalg}  (C22) & 0.113 & 0.137 & \textcolor{blue}{0.016} & 0.016 & 0.121 & 0.161 & \textcolor{blue}{0.043} & 0.052 & \textcolor{blue}{0.044} & 0.056 & \textcolor{blue}{0.041} & 0.086 & \textcolor{blue}{0.040} & 0.091 \\
\midrule
\midrule
NG (TSF) & 0.967 & 0.096 & 0.932 & 0.169 & 0.963 & 0.061 & -- & -- & -- & -- & -- & -- & 0.667 & 0.470 \\
w-CF (TSF) & -- & -- & -- & -- & -- & -- & -- & -- & \textcolor{blue}{0.006} & 0.000 & 0.818 & 0.385 & 0.207 & 0.397 \\
\midrule
AB-CF (TSF) & 0.589 & 0.218 & 0.871 & 0.030 & 0.307 & 0.063 & 0.410 & 0.284 & 0.580 & 0.298 & 0.589 & 0.265 & 0.528 & 0.324 \\
TSEvo (TSF) & 0.154 & 0.105 & 0.180 & 0.072 & 0.170 & 0.084 & 0.194 & 0.076 & 0.140 & 0.072 & 0.512 & 0.338 & 0.128 & \textcolor{blue}{0.040} \\
\midrule
\textbf{\newalg}  (TSF) & \textcolor{blue}{0.113} & 0.131 & \textcolor{blue}{0.054} & 0.037 & \textcolor{blue}{0.051} & 0.046 & \textcolor{blue}{0.049} & 0.029 & 0.030 & 0.027 & \textcolor{blue}{0.053} & 0.081 & \textcolor{blue}{0.033} & 0.044 \\
\bottomrule
\end{tabular}
}
\caption{The \textbf{Sparsity} metric for each counterfactual XAI method on two different classifiers (TSF and C22) and seven different benchmark datasets.\label{tab:sparsity}}
\end{table*}

\begin{table*}[!htbp]
\centering
\resizebox{\textwidth}{!}{
\begin{tabular}{l|*{8}{c@{\hskip 10pt}c}}
\toprule
\multirow{2}{*}{Explainer} & \multicolumn{2}{c}{ECG200} & \multicolumn{2}{c}{Coffee} & \multicolumn{2}{c}{GunPoint} & \multicolumn{2}{c}{CBF} & \multicolumn{2}{c}{Lightning7} & \multicolumn{2}{c}{ACSF1} & \multicolumn{2}{c}{Beef} \\
                           & Mean & Std. & Mean & Std. & Mean & Std. & Mean & Std. & Mean & Std. & Mean & Std. & Mean & Std. \\
\midrule
NG (C22) & 0.121 & 0.086 & 0.098 & 0.260 & 0.010 & 0.011 & 0.383 & 0.220 & -- & -- & -- & -- & 1.000 & 0.000 \\
w-CF (C22) & \textcolor{blue}{0.001} & 0.000 & -- & -- & \textcolor{blue}{0.001} & 0.000 & \textcolor{blue}{0.001} & 0.001 & -- & -- & \textcolor{blue}{0.00003} & 0.00002 & \textcolor{blue}{0.00003} & 0.0002 \\
\midrule
AB-CF (C22) & 0.132 & 0.081 & 0.024 & 0.007 & 0.069 & 0.059 & 0.236 & 0.144 & 0.222 & 0.114 & 0.060 & 0.084 & 0.103 & 0.195 \\
TSEvo (C22) & 0.311 & 0.161 & 0.021 & 0.011 & 0.036 & 0.023 & 0.275 & 0.097 & 0.240 & 0.139 & 0.090 & 0.075 & 0.036 & \textcolor{blue}{0.023} \\
\midrule
\textbf{\newalg}  (C22) & 0.062 & 0.076 & \textcolor{blue}{0.002} & 0.002 & 0.039 & 0.067 & 0.032 & 0.037 & \textcolor{blue}{0.045} & 0.062 & 0.008 & 0.017 & 0.047 & 0.133 \\
\midrule
\midrule
NG (TSF) & 0.164 & 0.118 & 0.026 & 0.011 & 0.062 & 0.029 & -- & -- & -- & -- & -- & -- & 0.515 & 0.365 \\
w-CF (TSF) & -- & -- & -- & -- & -- & -- & -- & -- & \textcolor{blue}{0.0001} & 0.0000 & 0.035 & 0.061 & \textcolor{blue}{0.0003} & 0.0001 \\
\midrule
AB-CF (TSF) & 0.161 & 0.067 & 0.045 & 0.009 & 0.043 & 0.037 & 0.168 & 0.110 & 0.251 & 0.169 & 0.065 & 0.073 & 0.092 & 0.150 \\
TSEvo (TSF) & 0.067 & 0.054 & 0.016 & 0.005 & 0.015 & 0.012 & 0.073 & 0.029 & 0.083 & 0.041 & 0.067 & 0.074 & 0.143 & 0.110 \\
\midrule
\textbf{\newalg}  (TSF) & \textcolor{blue}{0.056} & 0.060 & \textcolor{blue}{0.007} & 0.004 & \textcolor{blue}{0.012} & 0.019 & \textcolor{blue}{0.030} & 0.020 & 0.039 & 0.062 & \textcolor{blue}{0.012} & 0.020 & 0.020 & 0.036 \\
\bottomrule
\end{tabular}
}
\caption{The \textbf{L1-Proximity} metric for each counterfactual XAI method on two different classifiers (TSF and C22) and seven different benchmark datasets.\label{tab:l1prox}}
\end{table*}

\begin{table*}[!htbp]
\centering
\resizebox{\textwidth}{!}{
\begin{tabular}{l|*{8}{c@{\hskip 10pt}c}}
\toprule
\multirow{2}{*}{Explainer} & \multicolumn{2}{c}{ECG200} & \multicolumn{2}{c}{Coffee} & \multicolumn{2}{c}{GunPoint} & \multicolumn{2}{c}{CBF} & \multicolumn{2}{c}{Lightning7} & \multicolumn{2}{c}{ACSF1} & \multicolumn{2}{c}{Beef} \\
                           & Mean & Std. & Mean & Std. & Mean & Std. & Mean & Std. & Mean & Std. & Mean & Std. & Mean & Std. \\
\midrule
NG (C22) & 0.182 & 0.068 & 0.105 & 0.259 & 0.046 & 0.034 & 0.478 & 0.216 & -- & -- & -- & -- & 1.000 & 0.000 \\
w-CF (C22) & \textcolor{blue}{0.006} & 0.001 & -- & -- & \textcolor{blue}{0.008} & 0.000 & \textcolor{blue}{0.007} & 0.005 & -- & -- & \textcolor{blue}{0.001} & 0.000 & \textcolor{blue}{0.004} & 0.002 \\
\midrule
AB-CF (C22) & 0.193 & 0.086 & 0.041 & 0.008 & 0.118 & 0.083 & 0.349 & 0.130 & 0.376 & 0.127 & 0.145 & 0.167 & 0.132 & 0.228 \\
TSEvo (C22) & 0.440 & 0.145 & 0.045 & 0.011 & 0.096 & 0.047 & 0.430 & 0.067 & 0.430 & 0.161 & 0.200 & 0.155 & 0.045 & 0.031 \\
\midrule
\textbf{\newalg}  (C22) & 0.134 & 0.105 & \textcolor{blue}{0.017} & 0.008 & 0.076 & 0.098 & 0.121 & 0.076 & \textcolor{blue}{0.117} & 0.127 & 0.036 & 0.061 & 0.084 & 0.207 \\
\midrule
\midrule
NG (TSF) & 0.175 & 0.113 & 0.036 & 0.016 & 0.091 & 0.046 & -- & -- & -- & -- & -- & -- & 0.561 & 0.395 \\
w-CF (TSF) & -- & -- & -- & -- & -- & -- & \textcolor{blue}{0.001} & 0.000 & \textcolor{blue}{0.058} & 0.128 & 0.058 & 0.128 & \textcolor{blue}{0.002} & 0.000 \\
\midrule
AB-CF (TSF) & 0.209 & 0.068 & 0.055 & 0.010 & 0.101 & 0.090 & 0.283 & 0.094 & 0.332 & 0.160 & 0.137 & 0.149 & 0.111 & 0.172 \\
TSEvo (TSF) & 0.169 & 0.094 & 0.043 & 0.007 & 0.051 & 0.038 & 0.181 & 0.046 & 0.210 & 0.113 & 0.125 & 0.105 & 0.357 & 0.255 \\
\midrule
\textbf{\newalg} (TSF) & \textcolor{blue}{0.133} & 0.107 & \textcolor{blue}{0.027} & 0.011 & \textcolor{blue}{0.047} & 0.057 & 0.112 & 0.062 & 0.100 & 0.102 & \textcolor{blue}{0.045} & 0.060 & 0.060 & 0.104 \\
\bottomrule
\end{tabular}
}
\caption{The \textbf{L2-Proximity} metric for each counterfactual XAI method on two different classifiers (TSF and C22) and seven different benchmark datasets.\label{tab:l2prox}}
\end{table*}

\begin{table*}[!htbp]
\centering
\resizebox{\textwidth}{!}{
\begin{tabular}{l|*{8}{c@{\hskip 10pt}c}}
\toprule
\multirow{2}{*}{Explainer} & \multicolumn{2}{c}{ECG200} & \multicolumn{2}{c}{Coffee} & \multicolumn{2}{c}{GunPoint} & \multicolumn{2}{c}{CBF} & \multicolumn{2}{c}{Lightning7} & \multicolumn{2}{c}{ACSF1} & \multicolumn{2}{c}{Beef} \\
                           & Mean & Std. & Mean & Std. & Mean & Std. & Mean & Std. & Mean & Std. & Mean & Std. & Mean & Std. \\
\midrule
\midrule
\textbf{\newalg} (C22) & 6.95 & 2.73 & 7.46 & 3.52 & 10.70 & 4.08 & 14.45 & 4.39 & 25.07 & 8.93 & 24.86 & 8.49 & 20.80 & 6.60 \\
\midrule
\midrule
\textbf{\newalg} (TSF) & 4.89 & 0.92 & 4.79 & 0.82 & 4.12 & 1.26 & 5.27 & 1.00 & 5.89 & 1.67 & 6.58 & 2.28 & 6.13 & 1.28 \\
\bottomrule
\end{tabular}
}
\caption{The \textbf{Diversity} metric for each counterfactual XAI method on two different classifiers (TSF and C22) and seven different benchmark datasets. (For all other methods this is always $1.00$)\label{tab:diversity}}
\end{table*}




\begin{figure*}[htbp]
    \centering
    \begin{subfigure}{2\columnwidth}
        \centering
        \includegraphics[width=0.49\linewidth]{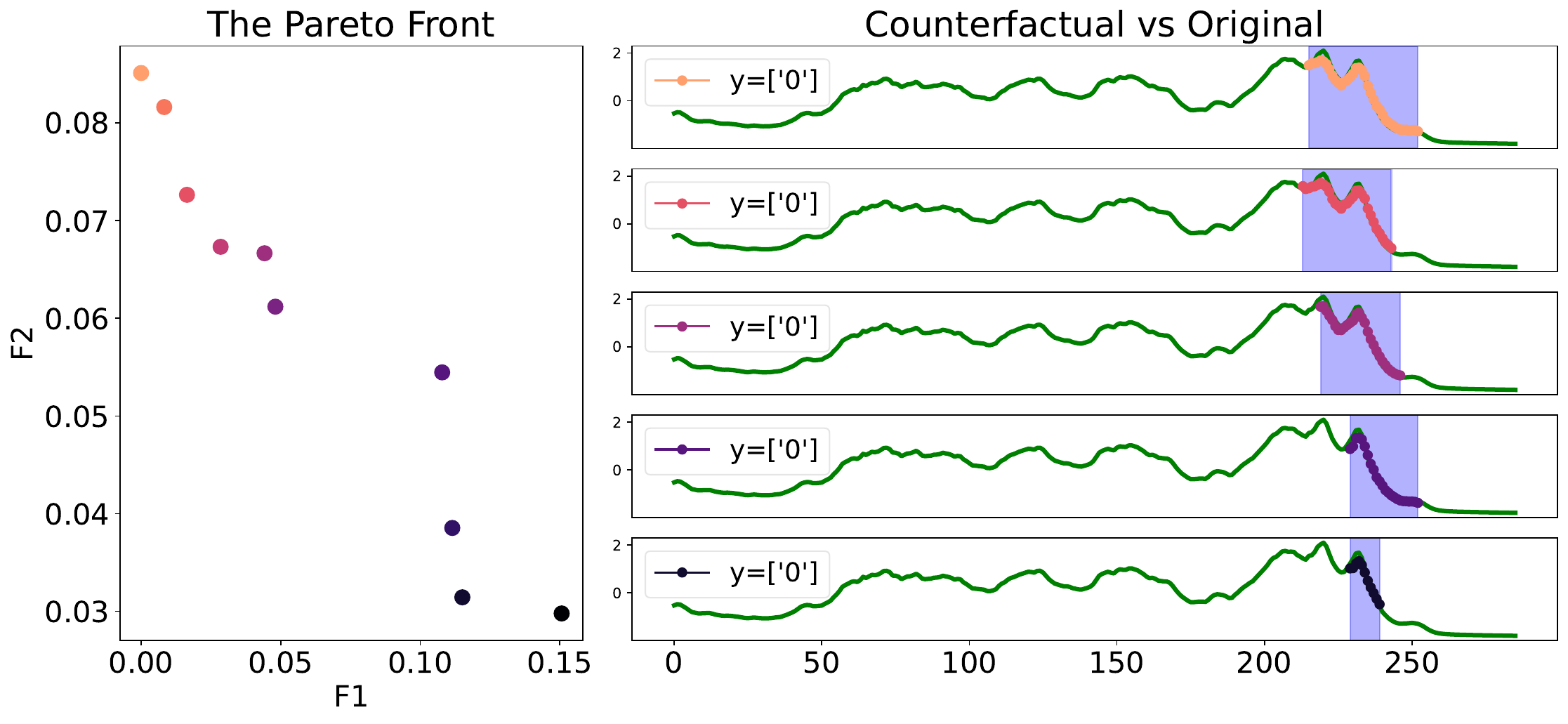}
        \includegraphics[width=0.49\linewidth]{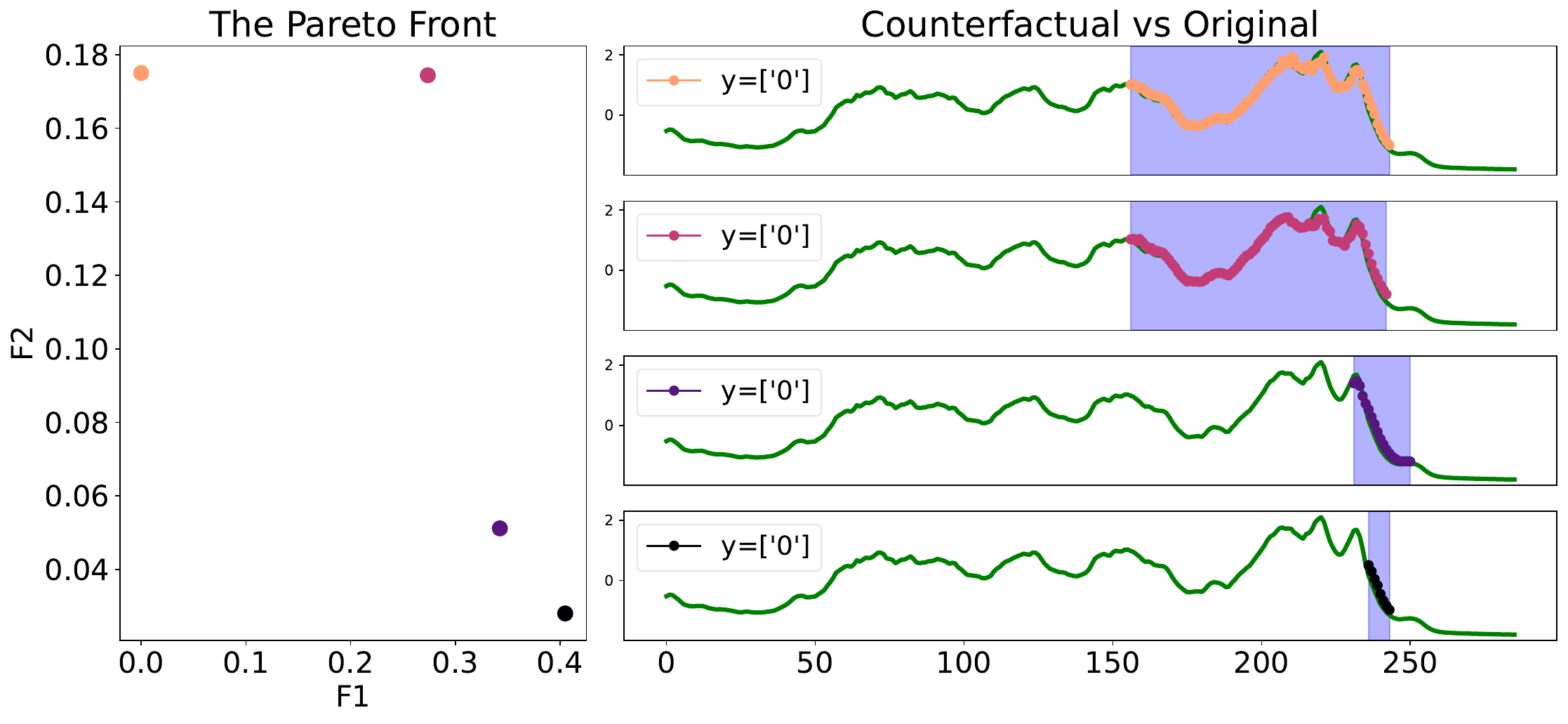}
        \caption{An example of the Pareto-efficient counterfactuals for the test sample (20) of a binary classification task \textit{Coffee} using Catch22 (left) and Supervised Time-series Forest (right).}
        \label{fig:example obj_coffee}
    \end{subfigure}\\
    \vspace{0.5cm}
    \begin{subfigure}{2\columnwidth}
        \centering
        \includegraphics[width=0.49\linewidth]{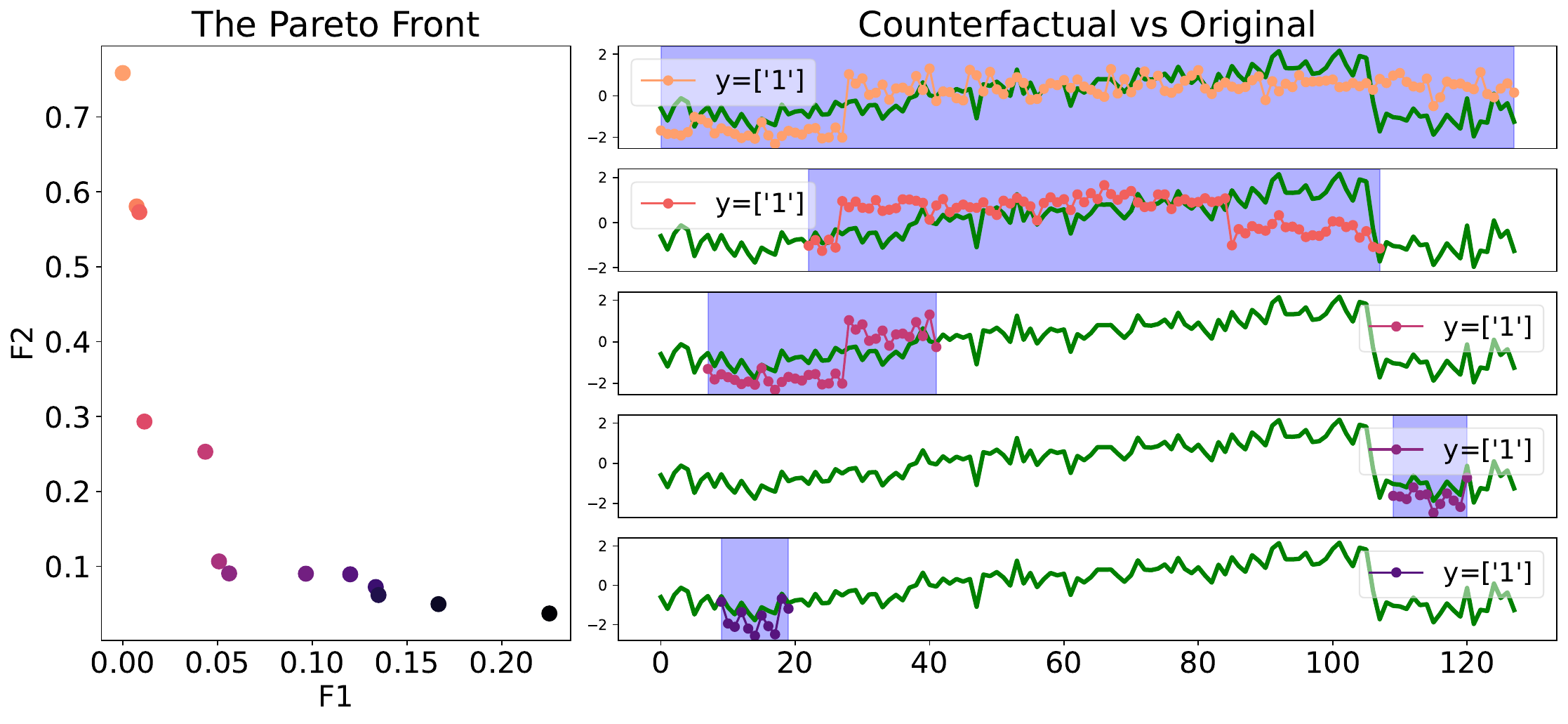}
        \includegraphics[width=0.49\linewidth]{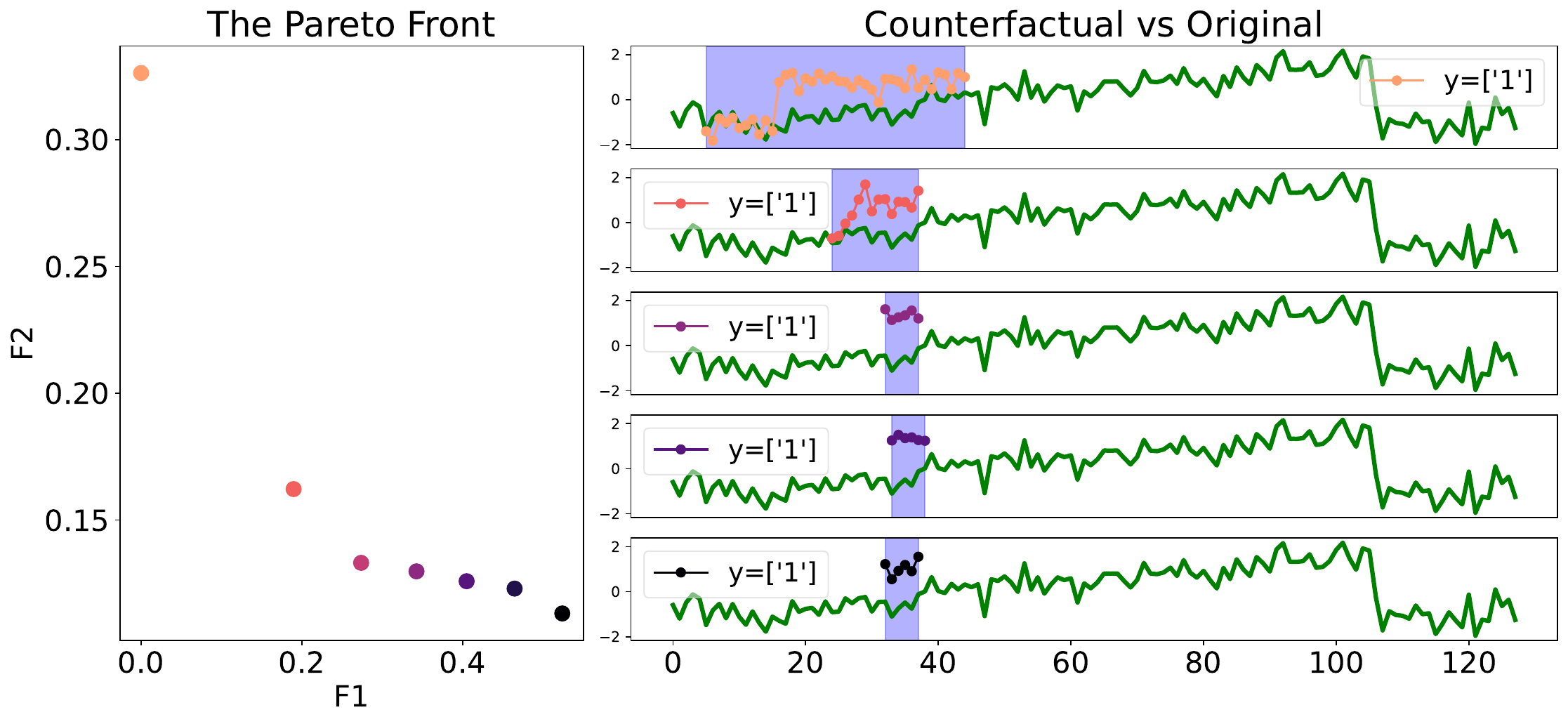}
        \caption{An example of the Pareto-efficient counterfactuals for the test sample (10) of a binary classification task \textit{CBF} using Catch22 (left) and Supervised Time-series Forest (right).}
        \label{fig:example obj_CBF}
    \end{subfigure}\\
    \vspace{0.5cm}
    \begin{subfigure}{2\columnwidth}
        \centering
        \includegraphics[width=0.49\linewidth]{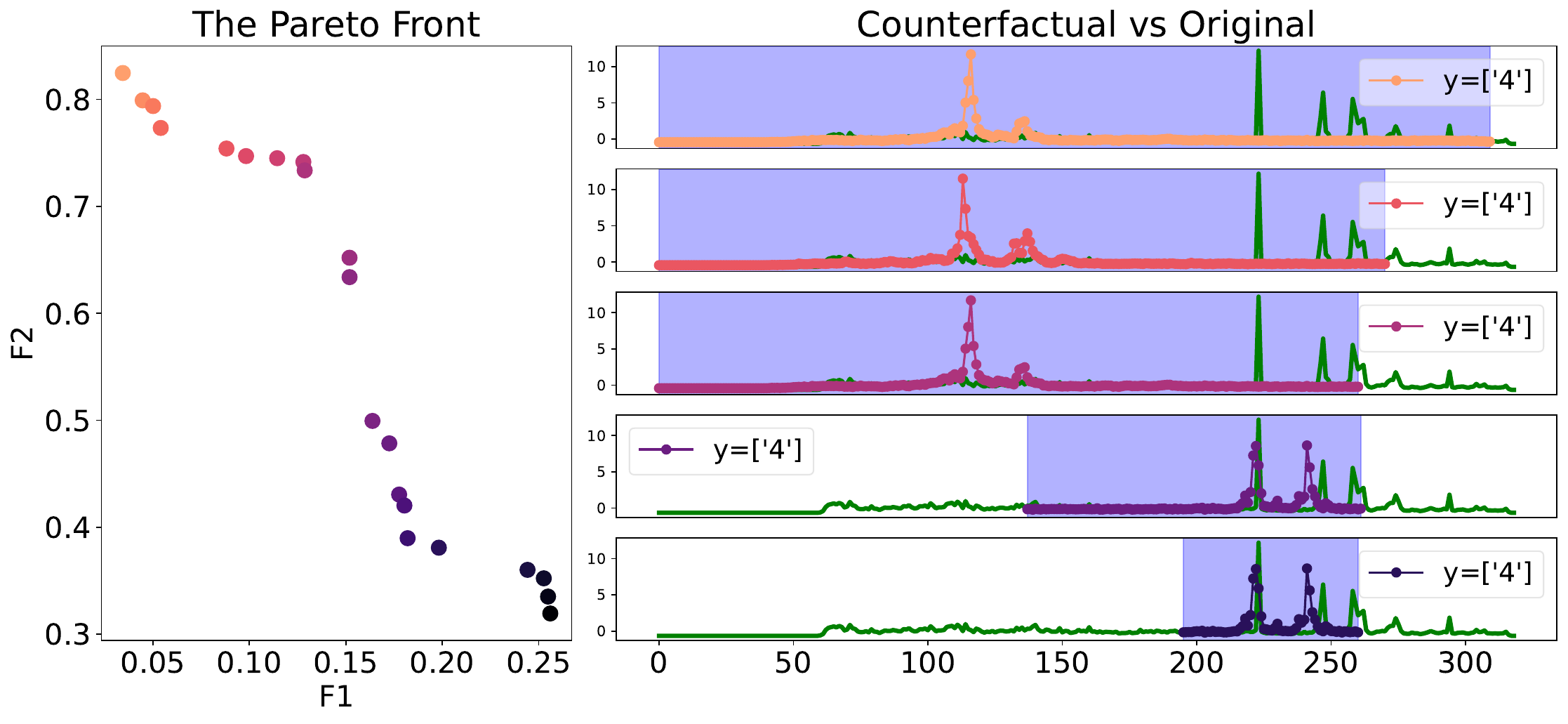}
        \includegraphics[width=0.49\linewidth]{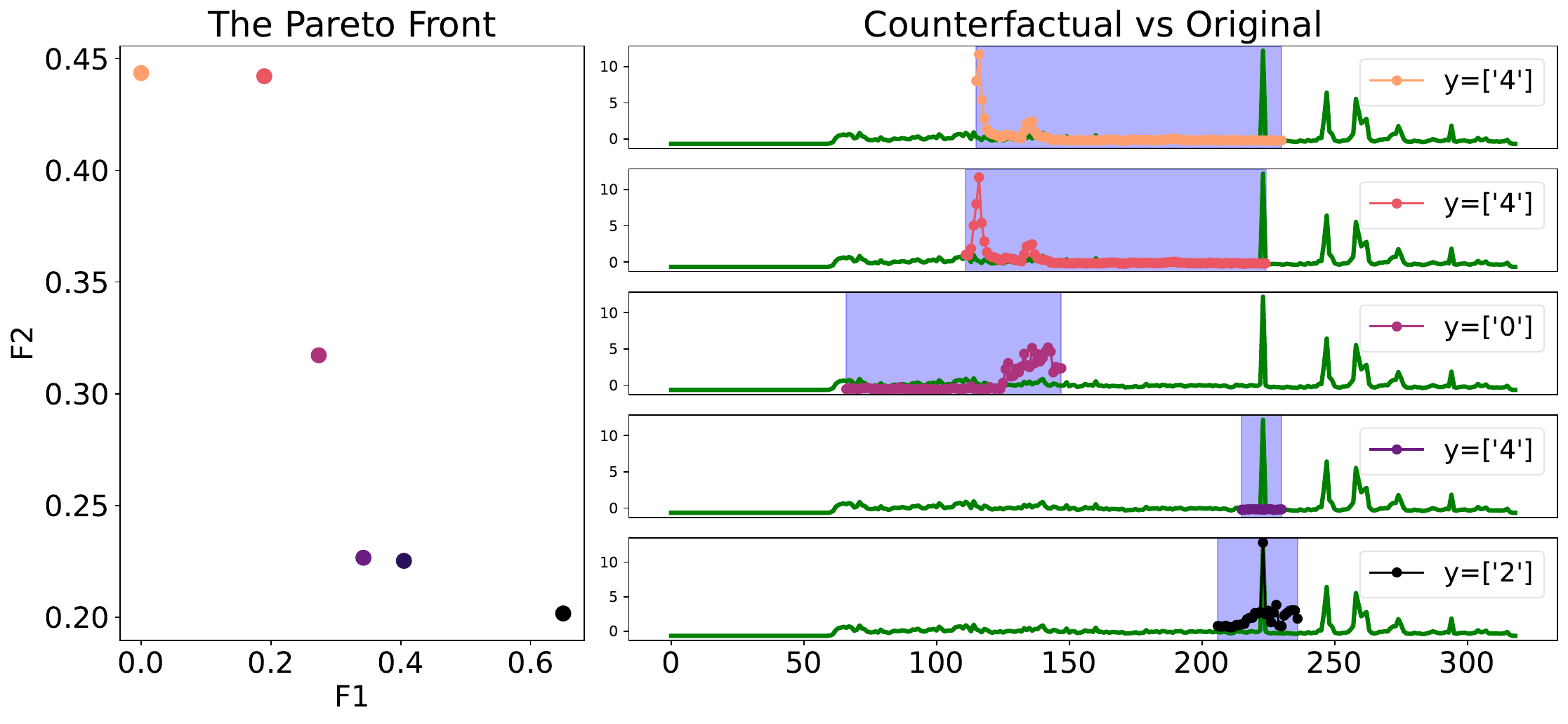}
        \caption{An example of the Pareto-efficient counterfactuals for the test sample (35) of a multi-class classification task \textit{Lightning7} using Catch22 (left) and Supervised Time-series Forest (right).}
        \label{fig:example obj_lightning7}
    \end{subfigure}\\
    \vspace{0.5cm}
    \begin{subfigure}{2\columnwidth}
        \centering
        \includegraphics[width=0.49\linewidth]{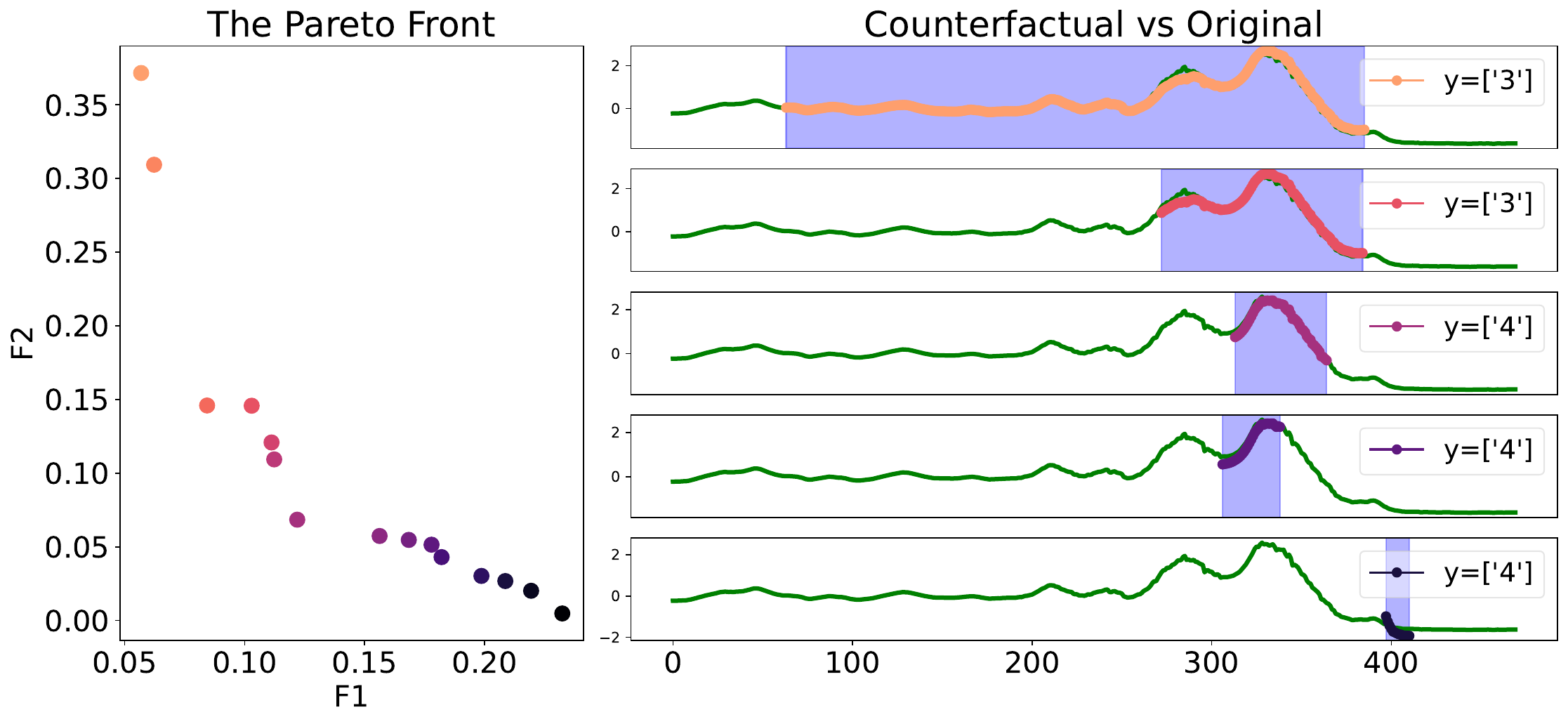}
        \includegraphics[width=0.49\linewidth]{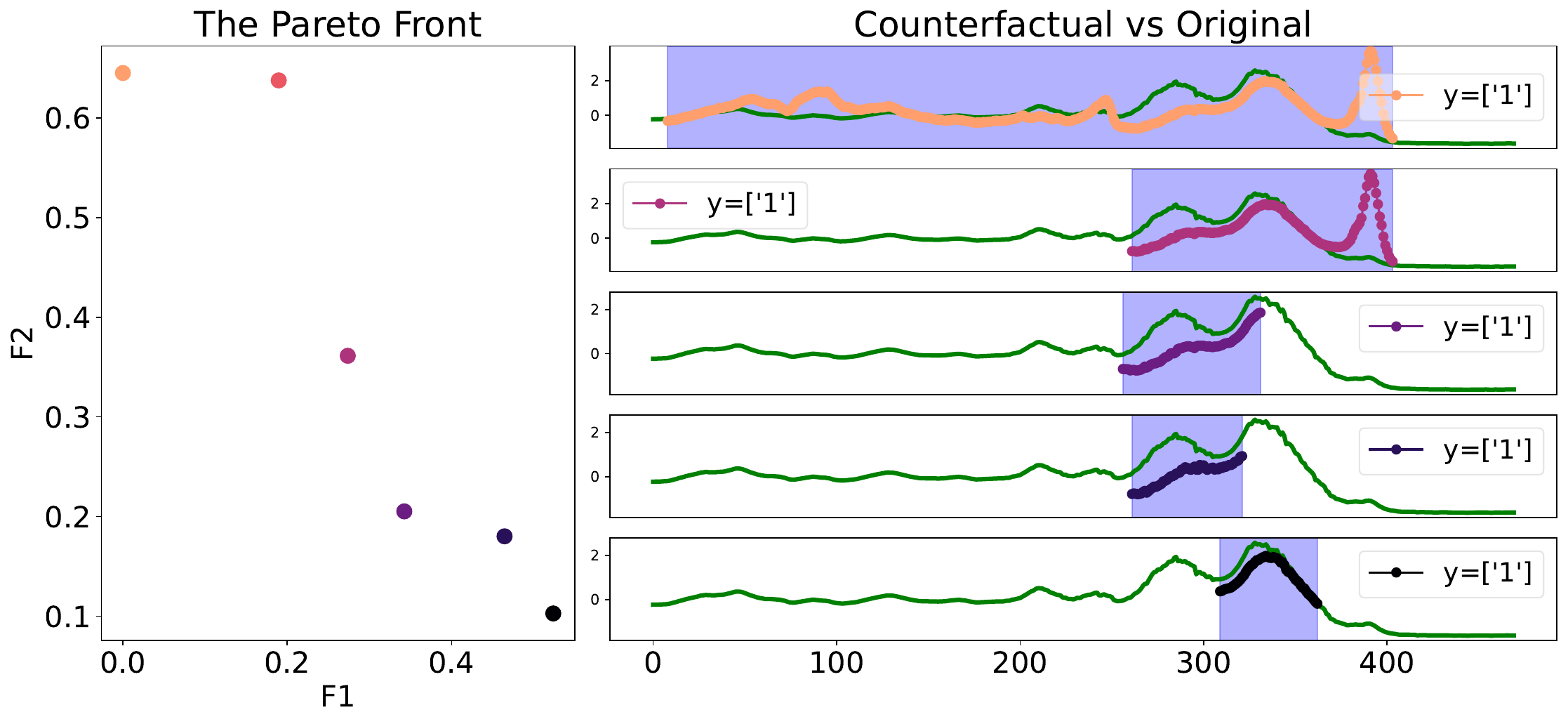}
        \caption{An example of the Pareto-efficient counterfactuals for the test sample (10) of a multi-class classification task \textit{Beef} using Catch22 (left) and Supervised Time-series Forest (right).}
        \label{fig:example obj_beef}
    \end{subfigure}
    
    \caption{
    Selection of examples of \newalg\ on different datasets and using different classifiers. In each subfigure, the left part shows the distribution of objective values (\(F_1\) on the \texttt{x} axis and \(F_2\) on the \texttt{y} axis) of the Pareto front. The right part of each figure displays the counterfactual examples and corresponding labels. The time series being explained is highlighted in~\textcolor{ForestGreen}{green}, while the counterfactual subsequence of interest (SoIs) are color-coded according to their position on the Pareto front.}\label{fig:exampleoptimizationresults}
\end{figure*}

\section{\uppercase{Results and Discussions}}

Our experimental results clearly demonstrate that \newalg\ consistently outperforms competing methods across several key metrics, as shown in Tables \ref{tab:validity}-\ref{tab:l2prox}. In terms of \emph{validity}, \newalg\ achieves a $100\%$ success rate in generating valid counterfactuals across all datasets, regardless of the classifier used (Table \ref{tab:validity}). This significantly outperforms baseline methods like w-CF, which struggles with generating valid counterfactuals, achieving a mere $1\% \sim 17\%$ validity across different datasets and classifiers. TSEvo, while better than w-CF, also lags behind \newalg\ in validity, particularly when using the Catch22 classifier. AB-CF is in terms of validity only marginally worse than our proposed solution.

\emph{Sparsity} is a key strength of \newalg, as demonstrated in Table \ref{tab:sparsity}. While methods like TSEvo also optimize for sparsity, \newalg\ achieves lower sparsity values across most datasets. 
In terms of proximity, Tables \ref{tab:l1prox} and \ref{tab:l2prox} show that \newalg\  performs competitively with state-of-the-art methods such as AB-CF and TSEvo, which also optimize for proximity. However, \newalg's use of multi-objective optimization allows it to maintain a strong proximity score while simultaneously achieving higher validity and diversity. The L1-Proximity and L2-Proximity results reflect that \newalg\ generates counterfactuals that are closer to the original time series, ensuring more interpretable and actionable insights.

As shown in the examples of Figure \ref{fig:exampleoptimizationresults} and also in Table \ref{tab:diversity}, our method generates a diverse set of Pareto-optimal solutions allowing for a wide range of plausible counterfactuals. All other methods only provide one counterfactual example per instance. This is particularly important in real-world applications, where generating multiple plausible alternatives can provide more comprehensive insights for end-users. The number of sub-sequences of the counterfactual examples by \newalg\ is always $1$, while for TSEvo this is on average $13.8$ sub-sequences, Native-Guide provides $1.8$, w-CF $1.7$ and AB-CF $1.8$ sub-sequences on average over all datasets. In general, less sub-sequences are more informative for counterfactual examples. However, it could be a limitation that the proposed method only provides $1$ sub-sequence, however, \newalg\ provides multiple examples per instance, one could combine the sub-sequences from these examples to alleviate this limitation. All results and source code can be found in our Zenodo repository\footnote{\url{https://doi.org/10.5281/zenodo.13711886}}.

In summary, \newalg's use of evolutionary multi-objective optimization and its reference-guided mechanism make it more effective than existing methods across key metrics. The method not only generates more valid and sparse counterfactuals but does so with a high degree of proximity and diversity, offering a significant advancement in the generation of explainable counterfactuals for time-series classification.

\section{\uppercase{Conclusions}}

In this paper, we presented \newalg, a novel framework for generating counterfactual explanations for time-series classification tasks using evolutionary multi-objective optimization. Through extensive experimentation, we demonstrated that \newalg\ consistently outperforms state-of-the-art methods in terms of validity, sparsity, proximity, and diversity across multiple benchmark datasets. By leveraging the flexibility of the NSGA-II algorithm and incorporating a reference-guided mechanism, our approach ensures that the generated counterfactuals are both interpretable and computationally efficient.

The good performance of \newalg\, particularly in achieving 100\% validity and maintaining high diversity while optimizing for proximity and sparsity, highlights its potential for real-world applications where model transparency is critical. Our proposed method strikes an effective balance between multiple conflicting objectives, offering a robust solution for generating meaningful counterfactuals in time-series classification.

Future work can explore several promising directions. First, further tuning of the hyper-parameters, particularly the number of reference instances, could lead to even greater improvements in performance. Additionally, extending \newalg\ to handle multivariate time-series data and real-time counterfactual generation could broaden its applicability to more complex, real-world scenarios. 

\FloatBarrier

\bibliographystyle{apalike}
{\small
\bibliography{main}}



\end{document}